\documentclass[runningheads]{llncs}

\usepackage{graphicx}
\usepackage{breakcites}
\usepackage{comment}
\usepackage{amsmath,amssymb}
\usepackage{color}
\usepackage{url}
\usepackage{hyperref}
\usepackage{tikz}
\newcommand{\tikzxmark}{%
\tikz[scale=0.23] {
    \draw[line width=0.7,line cap=round] (0,0) to [bend left=6] (1,1);
    \draw[line width=0.7,line cap=round] (0.2,0.95) to [bend right=3] (0.8,0.05);
}}
\newcommand{\tikzcmark}{%
\tikz[scale=0.23] {
    \draw[line width=0.7,line cap=round] (0.25,0) to [bend left=10] (1,1);
    \draw[line width=0.8,line cap=round] (0,0.35) to [bend right=1] (0.23,0);
}}

\begin{document}

% Replace with your title
\title{Self-supervised Vision Transformers for 3D Pose Estimation of Novel Objects}
% You can use \thanks for acknowledgment. Do not add any acknowledgment to the draft 
% version that is used for the review process.  
%\title{Title\thanks{XXX}}

\titlerunning{Self-supervised ViTs for Novel Object Pose}

\author{Stefan Thalhammer\inst{1}\orcidID{0000-0002-0008-430X} \and
Jean-Baptiste Weibel\inst{1}\orcidID{0000-0003-0201-4740} \and
Markus Vincze\inst{1}\orcidID{0000-0002-2799-491X}\and
Jose Garcia-Rodriguez\inst{2}\orcidID{0000-0002-7798-3055}}

\newcommand\blfootnote[1]{%
  \begingroup
  \renewcommand\thefootnote{}\footnote{#1}%
  \addtocounter{footnote}{-1}%
  \endgroup
}

\authorrunning{S. Thalhammer et al.}
	
\institute{Automation and Control Institute, TU Wien, Vienna, Austria
\email{\{thalhammer, weibel, vincze\}@acin.tuwien.ac.at}\\ 
\and 
Department of Computer Technology, University of Alicante, San Vicente del Raspeig, Spain\\
\email{jgarcia@dtic.ua.es}}

\maketitle              % typeset the header of the contribution

\begin{abstract}
Object pose estimation is important for object manipulation and scene understanding.
In order to improve the general applicability of pose estimators, recent research focuses on providing estimates for novel objects, that is objects unseen during training.
Such works use deep template matching strategies to retrieve the closest template connected to a query image.
This template retrieval implicitly provides object class and pose.
Despite the recent success and improvements of Vision Transformers over CNNs for many vision tasks, the state of the art uses CNN-based approaches for novel object pose estimation.
This work evaluates and demonstrates the differences between self-supervised CNNs and Vision Transformers for deep template matching.
In detail, both types of approaches are trained using contrastive learning to match training images against rendered templates of isolated objects. 
At test time, such templates are matched against query images of known and novel objects under challenging settings, such as clutter, occlusion and object symmetries, using masked cosine similarity. 
The presented results not only demonstrate that Vision Transformers improve in matching accuracy over CNNs, but also that for some cases pre-trained Vision Transformers do not need fine-tuning to do so. 
Furthermore, we highlight the differences in optimization and network architecture when comparing these two types of network for deep template matching.

\end{abstract}
\section{Introduction}
\label{sec:Introduction}

Object pose estimation is an important yet difficult vision problem.\blfootnote{source code: https://github.com/sThalham/TraM3D}
Many downstream tasks, such as grasping~\cite{patten2020dgcm}, augmented reality~\cite{hou2020mobilepose} and reconstruction~\cite{park2020neural} benefit from the availability of object poses.
%The complexity of this problem arises from deriving the $6D$ object pose from exclusively visual input. 
%Such poses are refereed to as 6D, which is a contraction for three degrees of freedom for translation and three for rotation. 
Classical object pose estimation approaches encode latent representations of multiple object views per object, during training.
During run-time these are matched against an observation to retrieve a coarse object pose~\cite{hinterstoisser2012model,hodavn2015detection,drost2010model}.
After retrieving the pose of the closest template, poses are refined using Iterative-Closest-Points~\cite{hartley2003multiple} algorithm or other algorithms to optimize the rigid transformations between two corresponding sets of points.
In contrast, learning-based solutions using Convolutional Neural Networks (CNN) learn a feature representation to infer object class and geometric correspondences during testing~\cite{Park_2019_ICCV,peng2019pvnet,thalhammer2021pyrapose,huang2022neural,thalhammer2023cope,wang2021gdr,aing2023faster,sun2022dynamic,dede2022object}. 
Yet, training pose estimators for each object instance~\cite{Park_2019_ICCV,peng2019pvnet}, or each set of object instances~\cite{thalhammer2023cope,wang2021gdr} is insufficient to be usable in real world scenarios where object instances are manifold and constantly changing.
As a consequence research shifts towards category-level~\cite{wang2019normalized,remus2023i2c-net} and novel object pose estimation~\cite{nguyen2022template,shugurov2022osop,labbe2022megapose}.
These recent novel object pose estimation approaches are similar to classical ones in the sense that queries are matched against templates.

The approach of~\cite{nguyen2022template} employs a CNN backbone to learn occlusion-aware template matching for novel object pose estimation.
Real observations are matched against rendered templates and tested for $3D$ pose estimation.
%, only requiring physicall-plausible rendering (pbr)~\cite{denninger} data for training.
While they show that such strategies are expedient for novel object pose estimation it has been shown that Vision Transformers (ViT)~\cite{dosovitskiy2020image,touvron2021training,caron2021emerging} learn more discriminative feature spaces than CNNs when trained in such unsupervised manners. 
This advantage of ViTs over CNNs, however, has primarily been empirically demonstrated by matching to distinct object classes and not by matching views of the same object class for more complex reasoning, such as $3D$ object pose estimation~\cite{caron2021emerging,chen2021empirical}.

In this work we empirically demonstrate that ViTs excel over CNNs when used for novel object pose estimation. 
Modifying the approach of~\cite{nguyen2022template} for comparing two similarly sized feature extractors, ResNet50~\cite{he2016deep} with $23M$ and ViT-s~\cite{touvron2021training} with $21M$ parameters, we show that these improvements are manifold.
Training self-supervised ViTs for $3D$ object pose estimation not only improves the template matching accuracy, but also reduces the training time.
Depending on the dataset and metric, template matching accuracy for seen objects ranges from $1\%$ on Linemod~\cite{hinterstoisser2012model}, over $4\%$ on Linemod-Occlusion~\cite{brachmann2014learning}, to $19\%$ on T-LESS~\cite{hodan2017tless}.
For unseen objects, the respective improvements are $3\%$, $5\%$ and $18\%$.
Achieving these improvements using ViT-s takes one fourth of the training time and iterations on LM and LM-O, and only one twenty-fifth of it on T-LESS.
More remarkably, testing ViT-s on T-LESS in a zero-shot fashion, thus without fine-tuning, already improves over using fine-tuned ResNet50 by $7\%$ and $9\%$, for seen and unseen objects respectively.
Finally, works such as~\cite{caron2021emerging,chen2021empirical} train self-supervised ViTs to retrieve the object class of seen objects assuming the availability of templates in the same domain.
These assumptions are impractical for novel object pose estimation.
Uniform coverage of the pose space is crucial and thus rendering templates is expedient. Furthermore, handling unseen objects is desired to further generalize real-world deployment of pose estimators.
As a consequence, this work provides ablations on the matter of network architecture used for matching.
While the aforementioned works~\cite{caron2021emerging,chen2021empirical} benefit from using high-dimensional, multi-layered projection heads, we empirically show that these increase the template matching error on unseen objects when matched against rendered templates. 
In summary we:
\begin{itemize}
    \item Show that Vision Transformers not only exhibit reduced template matching errors compared to CNNs for matching synthetic templates to known objects, but also to novel objects. The relative improvements for novel object pose estimation range from $3\%$ to $18\%$, depending on the dataset and metric used.
    \item Demonstrate that pre-trained Vision Transformers exhibit excellent matching performance for zero-shot matching. On the T-LESS dataset, non fine-tuned Vision Transformers exhibit a relative improvement over fine-tuned CNNs of $7\%$ and $9\%$, on known and novel objects respectively. Fine-tuning further improves to $19\%$ and $18\%$ respectively.
    \item Highlight the differences in matching procedure and optimization of fine-tuning Vision Transformers for template matching.
    Our results indicate that Vision Transformers encode relevant features over a broad range of descriptor sizes for seen and novel objects. As compared to CNNs, where there is a trade-off when choosing the descriptor size for either seen or novel objects. Our results additionally indicate that high-dimensional, multi-layered projection heads increase the template matching error for the problem at hand.
\end{itemize}

The remainder of the manuscript is organised in the sections Related Work, Method, Experiments and Conclusion.
The next section presents the state of the art for object pose estimation, focusing on deep template matching for deriving poses of novel objects, and self-supervised vision transformers.

\section{Related Work}
\label{sec:Related}

This sections presents the state of the art for object pose estimation with the focus on novel object pose estimation.
Subsequently, ViTs and self-supervised training for them is presented.

Learning-based object pose estimation research focuses on multi-staged pipelines~\cite{jiang2021triangulate,Park_2019_ICCV,wang2021gdr,sun2022dynamic} that often train separate networks for instance-level pose estimation~\cite{Park_2019_ICCV,wang2021gdr}, in order to improve the estimated pose's accuracy. 
Different streams of research improve on the scalability of instance-level pose estimation, presenting solutions for improved multi-object handling~\cite{aing2023faster,thalhammer2021pyrapose,zhang2020out} and reducing the number of stages needed for providing reliable pose estimates~\cite{thalhammer2023cope,dede2022object,zhang2019real}.
Yet, re-training pose estimators every time novel objects or object sets are encountered is cumbersome and delays the deployment in the real world.
As a consequence, recent works overcome these shortcomings by training for category-level pose estimation~\cite{wang2019normalized,remus2023i2c-net} or by training deep template matching for novel object pose estimation~\cite{nguyen2022template,shugurov2022osop,labbe2022megapose}.

\textbf{Deep Template Matching}
Matching observations against predefined templates is a long-standing concept of object pose estimation~\cite{hinterstoisser2012model,hodavn2015detection,drost2010model}.
Recent learning-based solutions adopt this strategy, since it has two major advantages~\cite{nguyen2022template,shugurov2022osop,labbe2022megapose}.
First, training time is low since encoding templates does not require learning a representation of each object individually.
Creating a latent representations for each relevant template only requires one network forward pass.
Thus, template encoding is done in the magnitude of seconds for an object of interest, as compared to training instance-level pose estimators, which takes hours to days, depending on the number of objects and the hardware~\cite{Park_2019_ICCV,wang2021gdr,thalhammer2023cope}.
Second, training instance-level pose estimators encodes a latent representation of the object, respectively objects, of interest.
This representation does not generalize to novel objects.
This shortcoming has to be addressed by either category-level object pose estimation, or by deep template matching.

The approach of~\cite{wohlhart2015learning} introduces deep descriptors for matching query objects against templates for retrieving the $3D$ pose using nearest neighbor search.
In~\cite{balntas2017pose} the authors improve over~\cite{wohlhart2015learning} by guiding learning in pose space, also accounting for object symmetries in the process.
Recently,~\cite{nguyen2022template} proposed further improvements.
They replace the triplet loss-based training with an InfoNCE-based one and improve occlusion handling by masking the feature embedding using the template's mask and an occlusion threshold.
We adopt and improve over their approach for deep template matching by using ViTs for descriptor extraction, which have not yet been adopted by the community. 
As such, we demonstrate their advantage with respect to their generality as deep template matcher and show empirical evaluations highlighting their advantages for the problem of novel object pose estimation.

\textbf{Vision Transformer}
It has recently been shown that ViTs~\cite{dosovitskiy2020image,parmar2018image} learn superior features when trained in a self-supervised fashion~\cite{chen2021empirical,caron2021emerging,touvron2021training}.
These mainstream works focus on training object classifiers from scratch, and using large datasets with little domain shift between query images and templates.
Such large datasets are difficult to obtain for object pose estimation due to the complexity of generation accurate $6D$ pose annotations. 
Additionally, it is relevant for pose estimation to effectively cover the viewing sphere around objects of interest~\cite{sundermeyer2023bop}.
This implies training on comparably small datasets and preferably using synthetically creating templates, i.e. using rendering for template creation~\cite{denninger}. 
As such, in this work, ViTs are assumed to be pre-trained, and templates are rendered.
We thus show the potential of self-supervised ViTs under that shifted perspective and also highlight the differences in network design as compared to the mainstream research direction.

\section{Method}
\label{sec:Method}

\begin{figure*}[th]
\caption{\textbf{Method overview} During training, a query image, a positive and a negative template is processed by a Vision Transformer to encode a feature embedding. The number of the positional tokens is retained for the feature map. InfoNCE~\cite{oord2018representation} is used in a Triplet loss-like fashion with the input feature map being masked with the positive template. During testing, novel query objects are matched against templates to retrieve object class and $3D$ pose from the matched template. Template retrieval is guided using the masked cosine similarity.}
\centering
\includegraphics[width=1.0\textwidth]{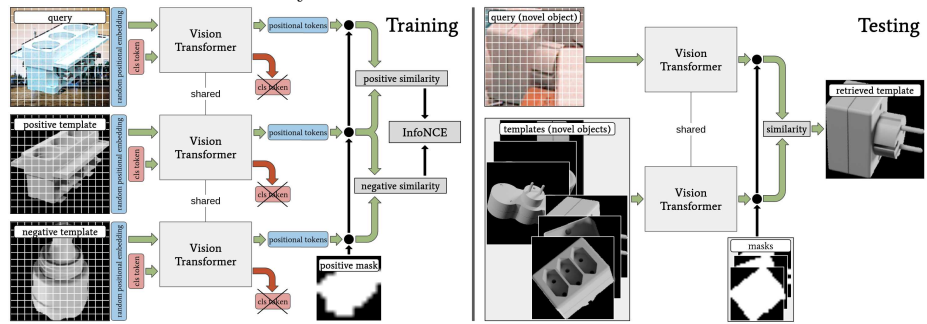}
\label{fig:approach_ov}
\end{figure*}

This section presents our self-supervised learning framework for matching real observations to synthetic templates for novel object $3D$ pose estimation.
Figure~\ref{fig:approach_ov} provides an abstract visualization of the presented method.

Self-supervised training is done using contrastive learning.
Contrastive learning aims at maximizing the similarity of semantically close training samples, referred to as positive pairs, while minimizing the similarity for samples that are semantically dissimilar, that is negative pairs.
More precisely, one training sample consists of a tuple of a query crop ($I_{q}$), a positive example ($I_{pos}$), and a negative one ($I_{neg}$).
The positive and negative template are rendered using physically-based rendering (pbr)~\cite{denninger}.
Where the positive sample correlates with respect to object class and rotation with the query image. 
%All input data is created by rendering using PBR.
The negative sample deviates with respect to both properties.
Crops are tokenized using random patch embedding and a shared pre-trained ViT-s~\cite{touvron2021training} is used for extracting features of the query and the template images.
In contrast to self-supervised ViT-frameworks for classification~\cite{caron2021emerging,chen2021empirical} we discard the class token and employ the positional tokens for similarity calculation. 
Using such spatial output enables dropping tokens based on the positive template's mask.
Optimization is guided using InfoNCE-loss~\cite{oord2018representation} with the positive and negative similarities as input.
During testing, similarities are computed between real object observations and pbr-templates of seen and novel objects.
Thus, in contrast to contemporary ViT-research, similarities have to bridge the synthetic-to-real gap, since templates are created using rendering~\cite{caron2021emerging,chen2021empirical}.
The real observations are compared against templates that represent uniformly distributed object views of the potentially new objects.
Ultimately, the class and the $3D$ rotation of the matched template are retrieved.

\subsection{Feature Embedding}
The aim of this work is novel object pose estimation.
Recent works shows that deep contrastively-learned template matching strategies are well suited for this task~\cite{nguyen2022template,shugurov2022osop,labbe2022megapose}.
In order to exhibit high similarities between similar view points of the same object in different domains, the learned feature embedding has to represent the object view as accurately as possible.
It has been shown that Vision Transformers~\cite{touvron2021training,dosovitskiy2020image,parmar2018image}, trained in an unsupervised way, learn to accurately model long-range image relationships, improving over CNNs~\cite{caron2021emerging}.

This works adopts the ViT-s network, presented in~\cite{touvron2021training} as feature extractor. 
The weights are pre-trained on ImageNet~\cite{krizhevsky2017imagenet} in a self-supervised manner~\cite{caron2021emerging}.
ViT-s is used by~\cite{caron2021emerging} only retaining the class token for training and testing.
In this work, the class token is discarded and the positional tokens are retained in order to benefit from the spatial nature of the output.
Diverse works indicate that augmenting feature extractors with deep multi-layered heads, for projecting embeddings to higher dimensions, improves performance when training on ImageNet~\cite{chen2021empirical,caron2021emerging,chen2020simple,grill2020bootstrap}.
The presented results in Section~\ref{sec:Experiments} indicate this finding does not apply to pose estimation.
A single linearly-activated fully-connected layer projects the feature embedding, coming from the pre-trained backbone, to a lower dimensionality.
It has to be noted that this different behavior is connected to the difference in problem; a) the backbone is initialized with pre-trained weights, b) the problem at hand matches real observations against rendered templates and c) testing is partially done on novel objects, thus data unseen during training.
We hypothesize that using deeper heads overfit to the training data characteristics.

The authors of~\cite{chen2021empirical} note that randomly initialized patch embedding stabilizes training on ImageNet and thus improves classification accuracy.
Accordingly, the patch embedding layer is not updated during fine-tuning.
Results are provided in Section~\ref{sec:Experiments}.

\subsection{Contrastive Learning Framework}

The feature embeddings extracted using ViT-s are processed by a contrastive learning framework for learning to increase similarity between object crops of the same class and a similar viewpoint.
As similarity measure, the cosine similarity is employed:

\begin{equation} \label{eq:1}
sim(emb_{I_{q},t}, emb_{\ast,t}) = \frac{emb_{I_{q},t} \cdot emb_{\ast,t} } { \lVert emb_{I_{q},t} \rVert_{2} , \lVert emb_{\ast,t} \rVert_{2} ) } 
\end{equation}

Where $\ast$ is  either $I_{pos}$ or $I_{neg}$.
The similarity is computed locally and aggregated for locations indicated by the mask image:

\begin{equation}
sim_{pos/neg} = \sum_{t=1}^{T} sim \left( I_{q},\ast \right) \times M_t \begin{cases} 
    sim &\text{if $M_{t}==1$}, \\
    0 &\text{otherwise}  
    \end{cases} 
\end{equation}

Where $T$ refers to the number of feature map locations, i.e. the number of positional tokens.
The negative similarity is summed over all embedded tokens inside the template's object mask, while the positive similarity is computed globally with $M = 1^{ \text{size of } I_{q}}$.
Both similarities are used in a triplet loss fashion~\cite{chechik2010large} using InfoNCE loss~\cite{gutmann2010noise,oord2018representation}.
Each positive sample is compared against all negative samples in a batch, resulting in $B = \left(b \cdot b\right) - b$ negative samples per iteration.

\begin{equation}
L = - \sum_{i=1}^{b} \log \frac{\exp \frac{sim_{pos,i}}{\tau} } {\sum_{k=1}^{B} \frac{sim_{neg,k}}{\tau}  \forall i \neq k} 
\end{equation}

Where $\tau$ is a temperature parameter set to $0.1$.
For more details consult~\cite{nguyen2022template}.

\subsection{Template Matching}

During testing templates of seen and novel objects, are matched against the query image. 
Embeddings are created for the query crop and all templates.
The cosine similarity in Equation~\ref{eq:1} is reused, yet modified to:

%\begin{equation}
%\begin{multline}
%sim = \sum_{0}^{t} \begin{cases}
%    sim_{t} \; &\text{if $sim_{t} > \delta$}, \\ 
%    0\; &\text{otherwise}  
%    \end{cases} 
%\end{equation}

\begin{equation}
sim_{q} = \sum_{t=1}^{T} sim \left( I_{q},\ast \right) \times M \begin{cases}\!%
  \begin{aligned}[b] 
    sim\;  &\text{if $M_{t}==1$}, \\
    &\text{and $sim_{t} > \delta$},\\
    0\;  &\text{otherwise}
  \end{aligned}\\
\end{cases} 
\end{equation}

Where $\delta$ is a hyperparameter set to $0.2$, which is meant to increase robustness against occluded image regions, as introduced by~\cite{nguyen2022template}.
The class and $3D$ rotation of the template leading to the highest cumulative cosine similarity are retrieved.

%%%%%%%%%%%%%%%%%%%%%%%%%%%%%%%%%%%%%%%%%%%%%%
%%% Experiments
%%%%%%%%%%%%%%%%%%%%%%%%%%%%%%%%%%%%%%%%%%%%%
\section{Experiments}
\label{sec:Experiments}

Presented results compare the CNN-based baseline methods, like~\cite{nguyen2022template}, to our approach that uses ViT-s as feature extractor.
Additional results evaluate the generality of the self-supervised pre-trained ViT-s without fine-tuning, showing that even without fine-tuning the template matching error is low and even improves over the baseline method on T-LESS.  
Ultimately, we present diverse ablations that highlight the differences between ViT- and CNN-architectures for $3D$ pose estimation.
The experiments section is concluded by providing an ablation with respect to the projection head used for our approach, highlighting the fundamental difference that for the addressed problem shallow heads are beneficial, as compared to approaches used for classification on ImageNet~\cite{caron2021emerging,chen2021empirical,chen2020simple,grill2020bootstrap}.

\subsection{Experimental Setup}

In the following paragraphs data retrieval and processing is detailed.
Following that, template creation for matching is explained.
In order to evaluate the proposed approach, standard metrics from concurrent, conceptually similar approaches and presented.

\subsubsection{Datasets}

Results are provided on three standard datasets for object pose estimation, Linemod~\cite{hinterstoisser2012model} (LM), Linemod-Occlusion~\cite{brachmann2014learning} (LM-O), and T-LESS~\cite{hodan2017tless}.
These datasets are processed to provide crop-level data in order to evaluate template matching accuracy and compare against the baseline method.

\textbf{LM and LM-O}  
These are two of the most-used datasets for evaluating object pose estimation approaches. 
LM features $13$ objects.
For each object a set of $\approx1200$ scene-level images is available.
Annotations are only provided for the respective object, though each set contains multiple objects of the dataset in the cluttered background.
The main characteristics of the dataset are texture-poor objects of different geometry, sizes and colors.
Annotated object views exhibit virtually no occlusion.
As as consequence,~\cite{brachmann2014learning} created annotations for all $8$ dataset objects in the Benchvise's set, thus introducing LM-O as a test set specifically for strongly occluded object views. 

With respect to training and test we follow~\cite{nguyen2022template}, in order to provide a fair comparison. 
For evaluation on seen and unseen objects the LM-objects are partitioned into three sets, see Table~\ref{tab:splits}.
As training data, $90\%$ of LM images' per object set are used, and the remaining $10\%$ are used for testing.
As a consequence training images are without occlusion.
The images of LM-O are exclusively used for testing, yet for evaluation also split accordingly, into seen and unseen objects.
In order to evaluate on all objects, one split is used for testing on unseen objects, while the other two are used training. 

\begin{table}[htbp]
\begin{center}
\caption{\textbf{LM/LM-O object splits.} Two of the sets are used for training and testing on seen objects, while the third is used for testing on unseen objects, as done by~\cite{nguyen2022template}.}
\begin{tabular}{|c|c|}
\hline
Split & Objects \\ \hline
1 & Ape, Benchvise, Camera and Can \\
2 & Cat, Driller, Duck and Eggbox \\
3 & Glue, Holepuncher, Iron, Lamp and Phone \\ \hline 
\end{tabular}
%\end{tabularx}
\label{tab:splits}
\end{center}
\end{table}

\textbf{T-LESS}
On T-LESS we follow the protocol of~\cite{sundermeyer2020multi}.
Isolated object views of the object $1-18$ are used for training and are pasted on a randomly chosen image of SUN397~\cite{xiao2010sun}, using the cut-paste strategy~\cite{dwibedi2017cut}.
These $18$ objects are considered as seen objects.
The remaining objects, $19-30$, are used as novel ones.
Test images are cropped from the primesense test set.

\subsubsection{Template Generation}

In contrast to works that train self-supervised ViTs for image classification~\cite{caron2021emerging,chen2021empirical}, this work considers matching the closest template for viewpoint classification, thus for $3D$ pose retrieval.
The major difference is that templates uniformly distributed in the viewing sphere, respectively hemisphere, are required.
Which is not relevant to the workings of aforementioned works.
Consequently, templates to match against are created using physically-based rendering for the task at hand~\cite{denninger}.

\textbf{LM and LM-O} 
The training and test dataset for LM and LM-O are processed as done by~\cite{wohlhart2015learning} and~\cite{nguyen2022template}.
These works crop the images from the real dataset by omitting in-plane rotations.
Thus, effectively only considering azimuth and elevation as degrees of freedom.
Objects are cropped in a way that the image space at object distance projects $0.4$ by $0.4$ meters.
Thus, all objects appear at the same distance to the camera, independent of their size.
Furthermore, neither the LM nor LM-O training and test images show objects from the lower viewing hemisphere.
Due to these constraints $301$ templates are sufficient for training and testing on LM and LM-O.

\textbf{T-LESS}
For T-LESS objects are cropped in a way to tightly encapsulate the objects.
Additionally, objects appear in arbitrary views in the test set.
As a consequence $92,232$ templates are used for training and testing on T-LESS, as done by~\cite{nguyen2022template,sundermeyer2020multi}.

\subsubsection{Evaluation}

This section presents the metrics used in this work.
The approach of~\cite{nguyen2022template} introduces \textit{Acc15} for evaluating template matching accuracy and classification.
The \textit{VSD}-score, as proposed by~\cite{hodan2018bop} is a standard metric for evaluating $6D$ object pose estimation accuracy.
The following paragraphs provide detailed explanations how these metrics are used in this work.

\textbf{\textit{Acc15}}
This metric is introduced by~\cite{nguyen2022template}.
It represents the accumulated true positive rate for matched templates that are below $15\deg$ rotational error with respect to the object class and ground truth rotation of the query crop:

\begin{equation}
%\begin{multline}
Acc15 = \sum_{n=1}^{n} 
\begin{cases}\!%
  \begin{aligned}[b]
    1\; &\text{if $\arccos{\frac{R_{q}\times R_{t}}{\lVert R_{q} \rVert_{2}\cdot \lVert R_{t} \rVert_{2}}}<15\deg$} \\ 
    &\text{and $C_{q} == C_{t}$},\\
    0\; &\text{otherwise}  
  \end{aligned}\\
\end{cases} 
\end{equation}
%\end{multline}

Where $n$ refers to the number of query crops, $R_{q}$ and $R_{t}$ to the three-dimensional rotation vectors, and $C_{q}$ and $C_{t}$ to the object class of the queries' ground truth and the template, respectively.
Thus, matched templates with a rotation deviation of more than $15\deg$ from the ground truth, or which have a different class than the query image, are considered as false positives. 

\textbf{\textit{VSD}}
This metric has been proposed by~\cite{hodan2018bop}.
For each query object crop the deviation of the estimated pose $\hat{P}$ to the ground truth $P$ is projected to a scalar value using:
\begin{equation}
    e_{VSD} = \underset{p\in \hat{V} \cup V}{avg} \begin{cases} 
    0 &\text{if $p \in \hat{V} \cap V \wedge | \hat{D}(p) - D (p)| < \tau$ }, \\
    1 &\text{otherwise}
    \end{cases} \\
\end{equation}

where $\hat{V}$ and $V$ are sets of image pixels; $\hat{D}$ and $D$ are distance maps and $\tau$ is a misalignment tolerance with the standard value of $20mm$. Distance maps are rendered and compared to the distance map of the test image to derive $\hat{V}$ and $V$.
Since $\hat{P}$ and $P$ need to represent $6D$ poses, including the $3D$ translation, we need to raise estimates to $6D$, the strategy of~\cite{sundermeyer2018implicit,nguyen2022template} is adopted.
Using the bounding box of the observation $box_{obs}$, and that of the template $box_{tmp}$, the corresponding intrinsics $f_{obs}$ and $f_{tmp}$, and the template distance to the camera $z_{tmp}$, enables deriving the observed object's distance $\hat{z}_{obs}$:

\begin{equation}
\hat{z}_{obs} = z_{tmp} \cdot \frac{\lVert box_{tmp,x}^2 \cdot box_{tmp,y}^2 \rVert_{2}} {\lVert box_{obs,x}^2 \cdot box_{obs,y}^2 \rVert_{2}} \cdot \frac{f_{obs}} {f_{tmp}}
\end{equation}

Using $\hat{z}_{obs}$, the relative translation between the observation and template of the other two translation parameters are derived. Where $\bullet$ is a placeholder for $x$ and $y$:

\begin{equation}
\Delta \bullet_{obs} = \frac{(box_{obs,\bullet} - c_{obs,\bullet}) \cdot  \hat{z}_{obs}}{f_{obs,\bullet}} - \frac{(box_{tmp,\bullet} - c_{tmp,\bullet}) \cdot  \hat{z}_{tmp}}{f_{tmp,\bullet}} 
\end{equation}

The $3D$ translation vector is ultimately composed as $t_{obs} = \{x_{tmp}+\Delta x_{obs}, y_{tmp}+\Delta y_{obs}, \hat{z}_{obs} \}$.

The \textit{VSD}-score is then defined as:

\begin{equation}
VSD = \sum_{n=1}^{n} \frac{1}{n}\begin{cases} 
    1 &\text{$e_{VSD,n} < 0.3$ }, \\
    0 &\text{otherwise}
    \end{cases} \\
\end{equation}

where $n$ again refers to the number of the query sample in an evaluated test set.

\subsection{Implementation Details}
\label{sec:Implementation}

This sections outlines the base method for comparing ViT to CNN-based template matching.
Following that the training procedure and the network architecture are detailed.

\textbf{Baseline method} For demonstrating the difference of CNNs and ViTs for self-supervised matching of real query crops to synthetic templates the baseline method of~\cite{nguyen2022template} is modified.
In order to provide a fair comparison all results are generated comparing backbones with a similar number of trainable parameters, ResNet50~\cite{he2016deep} with $23M$ and ViT-s~\cite{touvron2021training} with $21M$ parameters, pre-trained in a self-supervised manner~\cite{caron2021emerging} on~\cite{russakovsky2015imagenet}.
The following paragraph details training procedure and optimization settings.

\textbf{Optimizer Setting} 
As optimizer AdamW~\cite{you2019large} is used.
The batch size is set to $16$, which is also the case for the reference method~\cite{nguyen2022template}.
The ViT networks are only trained for five epochs, as compared to the baseline, which is trained for $20$ epochs. 
The linear scaling rule $lr = lr_b \cdot batch\_size/256$~\cite{goyal2017accurate} is adopted for choosing the learning rate.
A grid search was used to determine the base learning rate ($lr_b$) of $2.5 \cdot 10^-5$.
No learning rate scheduling is used.
Cosine weight decay scheduling, starting at $0.04$ and ending at $0.4$ after two epochs, is employed.

The input image size is $224^2$ and the template's mask size $14^2$.
A patch size of $16$ is used for input image tokenization.
A single linear layer is used to project the backbone feature size of $384$ to $32$.
This stands in contrast to works like~\cite{caron2021emerging,chen2021empirical,chen2020simple}, where multi-layered high-dimensional projectors are used.
The input to the projection head is normalized using batch normalization~\cite{ioffe2015batch}.
The output of the projector is normalized using~\cite{ba2016layer}.
Section~\ref{sec:Ablations} ablates mask and descriptor size, as well as the choice for the projection head.

%\textbf{6D Pose Retrieval}
%Template Matching retrieves the closest template, thus, the $3D$ rotation of the query frame.
%In order to raise these estimates to $6D$, the strategy of~\cite{sundermeyer2018implicit,nguyen2022template} is adopted.
%Using the bounding box of the observation $box_{obs}$, and that of the template $box_{tmp}$, the corresponding intrinsics $f_{obs}$ and $f_{tmp}$, and the template distance to the camera $z_{tmp}$, enables deriving the observed object's distance $\hat{z}_{obs}$:

%\begin{equation}
%\hat{z}_{obs} = z_{tmp} \cdot \frac{\lVert box_{tmp,x}^2 \cdot box_{tmp,y}^2 \rVert_{2}} {\lVert box_{obs,x}^2 \cdot box_{obs,y}^2 \rVert_{2}} \cdot \frac{f_{obs}} {f_{tmp}}
%\end{equation}

%Using $\hat{z}_{obs}$, the relative translation between the observation and template of the other two translation parameters are derived. Where $\bullet$ is a placeholder for $x$ and $y$:

%\begin{equation}
%\Delta \bullet_{obs} = \frac{(box_{obs,\bullet} - c_{obs,\bullet}) \cdot  \hat{z}_{obs}}{f_{obs,\bullet}} - \frac{(box_{tmp,\bullet} - c_{tmp,\bullet}) \cdot  \hat{z}_{tmp}}{f_{tmp,\bullet}} 
%\end{equation}

%The $3D$ translation vector is ultimately composed as $t_{obs} = \{x_{tmp}+\Delta x_{obs}, y_{tmp}+\Delta y_{obs}, \hat{z}_{obs} \}$.

\subsection{Main Results}
This section presents experiments comparing ResNet50~\cite{he2016deep} as feature extractor to ViT-s~\cite{touvron2021training}.
Evaluations are provided comparing to the state of the art for $3D$ template matching to the presented approach.

\subsubsection{Results on LM/LM-O}

Table~\ref{tab:LM} compares the presented approach to those of~\cite{wohlhart2015learning}, \cite{balntas2017pose} and~\cite{nguyen2022template} for template matching on LM and LM-O.
Reported are the true positive rates of matched templates with respect to object class and rotational error below $15\deg$ (\textit{Acc$15$}), as defined in~\cite{wohlhart2015learning}.
We follow the paradigm of~\cite{nguyen2022template} and report the results of the best-performing epoch during fine-tuning.
The results show that using ViTs as feature extractor consistently outperforms the CNN approach for objects seen and unseen during training.
Both, conceptually similar approaches, use backbones with a comparable amount of parameters, ResNet50~\cite{he2016deep} with $23M$ and ViT-s~\cite{touvron2021training} with $21M$.
It has to be mentioned that the method of~\cite{nguyen2022template} is fine-tuned for $20$ epochs while the ViTs are fine-tuned for only $5$. 

\begin{table}[htbp]
\begin{center}
\caption{\textbf{Comparison on LM/LM-O.} Amount of true poses for a rotational error threshold of $15\deg$ (\textit{Acc15}~\cite{wohlhart2015learning}) for objects seen and unseen during training, see Table~\ref{tab:splits}. The compared backbones have similar parameters, $23M$ for ResNet50~\cite{he2016deep} and $21M$ for ViT-s~\cite{touvron2021training}. Results for the methods indicated with $\dagger$ are taken from~\cite{nguyen2022template}.}
\begin{tabular}{|c|c|c|c|c|c|}
%\begin{tabularx}{\columnwidth}{|c|c|c|c|c|c|c|}
\hline
\multicolumn{2}{|c|}{} & \multicolumn{2}{c|}{seen} & \multicolumn{2}{c|}{unseen} \\ 
%\cline{5-7}
%App. & Backbone & $\vartheta$ & \textbf{LM} & \textbf{LM-O} & \textbf{LM} & \textbf{LM-O} \\ \hline
Method & Backbone & LM & LM-O & LM & LM-O \\ \hline
\cite{wohlhart2015learning}$\dagger$ & RN50\cite{he2016deep} & 98.1 & 67.5 & 45.1 & 29.9 \\
\cite{balntas2017pose}$\dagger$ & RN50\cite{he2016deep}  & 96.1 & 64.7 & 44.3 & 29.1 \\
\cite{nguyen2022template} & RN50\cite{he2016deep}  & 99.1 & 79.4 & 93.5 & 76.3 \\
Ours & ViT-s\cite{touvron2021training} & \textbf{99.8} & \textbf{82.2} & \textbf{96.4} & \textbf{80.2} \\ \hline 
\end{tabular}
%\end{tabularx}
\label{tab:LM}
\end{center}
\end{table}

\begin{figure}[h]
\caption{\textbf{Results on LM and LM-O splits in detail.} Reported is the percentage of true poses for different rotational error thresholds, of the CNN- and ViT-backbone for the seen and unseen object splits.}
\centering
\includegraphics[width=1.0\columnwidth]{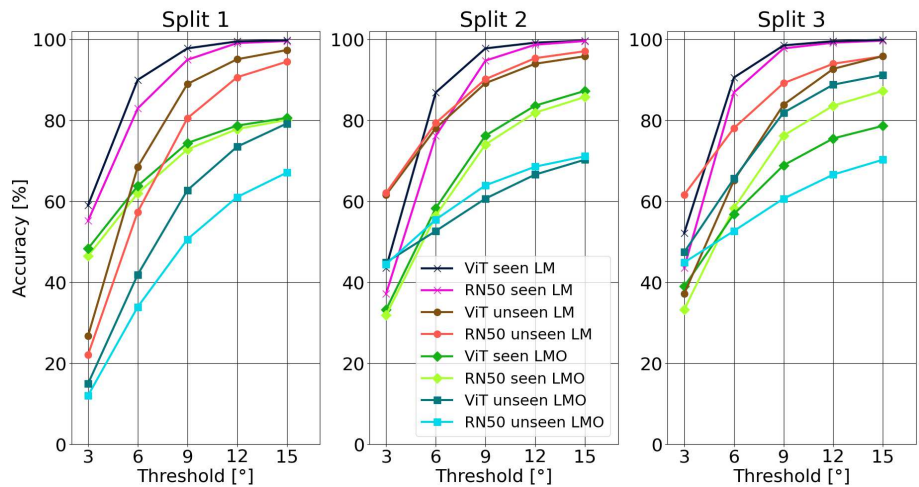}
\label{fig:LM}
\end{figure}

Figure~\ref{fig:LM} shows a detailed comparison for the individual data splits of LM and LM-O, using ResNet50~\cite{he2016deep} and ViT-small~\cite{touvron2021training} as feature extractors for template matching..
Tendentiously, ViT-s improves in pose estimation with respect to all rotational error thresholds on all the splits.
The only exceptions are the seen LM split 3, unseen LM-O split $2$ and seen LM-O split $3$. %This indicates that the larger mask size of CNNs ($32$ as compared to $14$ for ViT) is beneficial for highly-occluded cases, see Figure~\ref{tab:splits}.

\subsubsection{Results on T-LESS}

Table~\ref{tab:tless} compares the proposed approach to the approaches of~\cite{nguyen2022template}, \cite{sundermeyer2020multi} and \cite{sundermeyer2018implicit}.
We follow the evaluation paradigm of~\cite{sundermeyer2020multi} and report the VSD-score~\cite{hodan2018bop} using the standard thresholds, and the ground truth bounding box as basis for translation estimation.
We report the performance after one epoch of fine-tuning, as compared to the $25$ epochs for~\cite{nguyen2022template}.
The results show that our approach, using ViT-small~\cite{touvron2021training} as feature extractor, consistently outperforms the competing approaches for objects seen and unseen during training.
Especially relevant is the comparison to the conceptually similar approach of~\cite{nguyen2022template}, which again uses ResNet50~\cite{he2016deep} as backbone.
%The last row shows results when using the ViT-small, pre-trained on ImageNet~\cite{russakovsky2015imagenet}, without fine-tuning.
These results show that ViTs work well for industrial objects of T-LESS, resulting in similar pose estimation accuracy for seen and unseen objects.
The following section presents pose estimation results using ViTs without fine-tuning.

%\begin{table}[<options>]
%\caption{}\label{tbl1}
%\begin{tabular*}{\tblwidth}{@{}LL@{}}
%\toprule
%  &  \\ % Table header row
%\midrule
% & \\
% & \\
% & \\
% & \\
%\bottomrule
%\end{tabular*}
%\end{table}

%\newcolumntype{b}{X|}
%\newcolumntype{s}{|{\hsize=.6\hsize}X|}
%\newcolumntype{h}{{\hsize=1.3\hsize}X|}
%\newcommand{\heading}[1]{\multicolumn{1}{X|}{#1}}

\begin{table}[htbp]
\begin{center}
\caption{\textbf{Comparison on T-LESS.} Results are presented using the \textit{VSD}-score with the standard thresholds presented in~\cite{hodan2020epos}.}
\begin{tabular}{|c|c|c|c|}
%\begin{tabularx}{\columnwidth}{sbbb}
%\begin{tabularx}{\columnwidth}{sbhs}
\hline
%\multicolumn{1}{|c}{} & \multicolumn{3}{c|}{Recall \textit{VSD}} \\ 
%\cline{2-4}
Method  & \textbf{seen:} & \textbf{unseen:} & Average \\ 
& Objects 1-18 & Objects 19-30 &  \\ \hline
\cite{sundermeyer2018implicit} & 35.60 & 42.45 & 38.34 \\
\cite{sundermeyer2020multi} & 35.25 & 33.17 & 34.42 \\
\cite{nguyen2022template} & 59.62 & 57.75 & 58.87 \\ 
Ours & \textbf{70.65} & \textbf{68.03} & \textbf{69.71} \\ \hline 
%Ours (pretrained only) & 63.93 & 62.93 & 63.57 \\ \hline 
\end{tabular}
%\end{tabularx}
\label{tab:tless}
\end{center}
\end{table}

\subsection{Feature Extractor Fine-Tuning}
\label{sec:finetuning}

This section discusses and presents results using only ImageNet-pretrained ViTs as feature extractor.
In order to use the pre-trained backbone without fine-tuning, the last linear projection layer is discarded.
The output dimensionality per feature map location is $384$.
Table~\ref{tab:ft} compares the presented approach with and without fine-tuning (indicated with "f.t." in the table) on LM, LM-O and T-LESS. 
The pre-trained ViT-s demonstrate tremendous generality with respect to feature embedding.
On the LM and LM-O datasets the matching accuracy using \textit{Acc15} is higher than that of~\cite{wohlhart2015learning} and~\cite{balntas2017pose}, see Table~\ref{tab:LM}.
Yet, fine-tuning improves for all test cases.
The matching accuracy on both, seen and unseen, T-LESS sets, evaluated using the \textit{VSD}-metric, is higher than for all methods compared against in Table~\ref{tab:tless}, even without fine-tuning.
Fine-tuning further improves performance.
The presented evaluation shows that ViTs pre-trained in a self-supervised fashion learn features that translate well to new tasks with a large shift in object categories, even without fine-tuning.

\begin{table}[htbp]
\begin{center}
\caption{\textbf{Influence of fine-tuning.} Result comparison for fine-tuning (f.t.) the ViT-s backbone versus only using the pre-trained feature extractor without fine-tuning.}
\begin{tabular}{|c|c|c|c|c|c|}
%\begin{tabularx}{\columnwidth}{|X|X|X|X|X|X|}
\hline
Metric & f.t. & \multicolumn{2}{|c|}{\textbf{seen}} & \multicolumn{2}{|c|}{\textbf{unseen}} \\ \hline
$Acc15$~\cite{wohlhart2015learning} & & LM & LM-O & LM & LM-O \\ \hline
 & \tikzxmark & 81.3 & 56.3 & 85.1 & 63.6 \\
& \tikzcmark & \textbf{99.8} & \textbf{82.2} & \textbf{96.4} & \textbf{80.2} \\ \hline
\textit{VSD}~\cite{hodan2020epos} & & \multicolumn{4}{|c|}{T-LESS} \\ \hline
% & & \multicolumn{2}{|c|}{Obj. 1-18} & \multicolumn{2}{|c|}{Obj. 19-30} \\ \hline
& \tikzxmark & \multicolumn{2}{|c|}{63.93} & \multicolumn{2}{|c|}{62.93} \\ 
 & \tikzcmark & \multicolumn{2}{|c|}{\textbf{70.65}} & \multicolumn{2}{|c|}{\textbf{68.03}} \\ 
\hline 
\end{tabular}
%\end{tabularx}
\label{tab:ft}
\end{center}
\end{table}

\subsection{Ablation Study}
\label{sec:Ablations}

This sections discusses the difference in output space size and descriptor size for CNNs and ViTs.
ViT and CNN approaches benefit from multi-layered, high-dimensional projection heads~\cite{chen2021empirical,caron2021emerging,chen2020simple,grill2020bootstrap}.
Ultimately, we present experiments on the influence of projection head on our approach and additional architecture choices.

\begin{figure}[h]
\centering
\includegraphics[width=1.0\columnwidth]{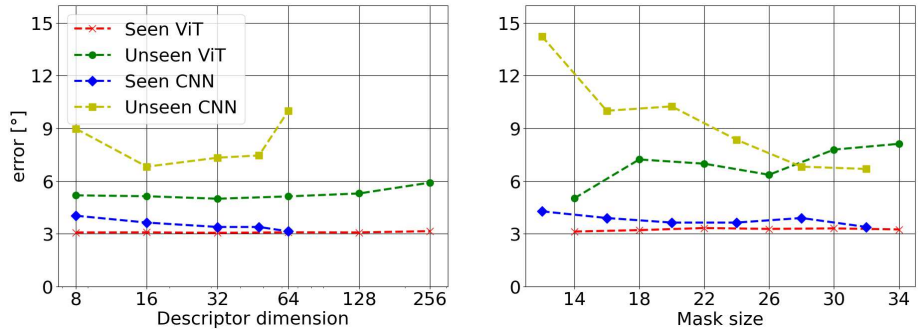}
\caption{\textbf{Influence of the descriptor and mask size on LM seen and unseen} The left plot shows the influence on the rotational error of the retrieved templates when using ResNet50 and ViT-s with different descriptor sizes. The mask size is set to $32$ for ResNet50 and to $14$ for ViT-s. The right plot show the same comparison for different mask sizes. The descriptor size is set to $16$ for ResNet50 and to $32$ for ViT-s.}
\label{fig:mask_desc}
\end{figure}

\subsubsection{Descriptor Size}

The left plot of Figure~\ref{fig:mask_desc} evaluates the influence of the descriptor size on the presented approach, and the ResNet50 baseline one on the seen and unseen sets of LM.
The cumulative rotational error on LM decreases steadily with increasing descriptor size when using ResNet50.
Yet, the optimal dimensionality is $16$ for minimizing the rotational error for the unseen LM objects.
While the descriptor size has a large influence on the seen LM set and even more on the unseen one, the behaviour using ViT-s is vastly different.
For ViT-s the descriptor dimensionality has little influence and leads to low errors over a broad range of dimensions for seen and unseen objects.
%Interestingly, the descriptor size that minimizes the error is different for both sets using ResNet50.
While for ResNet50 the error progression is different for both sets, the dimensionality that minimizes the error on both sets is $32$ when using ViT-s.

\subsubsection{Mask Size}

The matching accuracy of the baseline method~\cite{nguyen2022template} increases when using spatially higher-dimensional feature maps since occlusion handling improves.
In order to use larger feature maps for computing the template similarities we adopt the projection head of the baseline.
%In order to evaluate the influence of the mask size on the template matching accuracy we adopt a similar projector design as~\cite{nguyen2022template}. 
Instead of using two convolutional layers for downsampling, we employ two transposed convolutional layers for upsampling.
Both are ReLU~\cite{nair2010rectified}-activated.
The first projecting the $384$ dimensional feature vectors output by the backbone to $256$, the second one to $32$.
Both convolutions apply no feature map padding, slide with a stride of one over the feature map and use the same kernel size, which is set depending on the desired mask size to either $3$, $5$, $7$, $9$, or $11$.
This projector replaces the projection head detailed in Section~\ref{sec:Implementation}.

The right plot of Figure~\ref{fig:mask_desc} evaluates the influence of the mask size on the rotational error of the matched templates.
For the presented comparison the ResNet50 baseline approach~\cite{nguyen2022template} uses a descriptor size of $16$, and ViT-s is used with a descriptor size of $32$.
With the ResNet50 backbone, for both the the seen and unseen objects the rotational error reduces with increasing mask size.
Using the proposed ViT-s approach the behaviour is again vastly different.
While the influence of the mask size is negligible for the seen objects, the rotational error for the novel objects increases significantly when increasing the mask size.
This indicates that ViT-s learns relevant features for the seen objects during fine-tuning with projection heads with larger spatial output.
As such, the template matching accuracy remains constant.
However, increasing the feature map size used for matching is detrimental for novel objects.
This correlates with the results presented in Section~\ref{sec:finetuning}, which indicate that ViTs already generalize well without fine-tuning.
The feature projection learned by a projection head with increased spatial output is less general and thus increases template matching error for novel objects.

\subsubsection{Network Architecture Design}

\begin{table}[htbp]
\begin{center}
\caption{\textbf{Network architecture.} Reported is the average rotational error on LM and LM-O. The projection heads output a feature dimensionality of 32. When no head is used the standard ViT-s dimensionality of $384$ is output. The column patch embedding (p.e.) indicates if the patch embedding layer is updated (l) or frozen (r) during fine-tuning.}
\begin{tabular}{|c|c|c|c|c|c|c|}
\hline
\multicolumn{3}{|c|}{} & \multicolumn{2}{|c|}{seen} & \multicolumn{2}{|c|}{unseen} \\ \hline
\textbf{Head} & p.e. & act. & \textbf{LM} & \textbf{LM-O} & \textbf{LM} & \textbf{LM-O} \\ \hline 
none & l & & 3.14 & 10.95 & 7.80 & 15.44 \\ 
 &  r & & 3.27 & 10.96 & 5.87 & 13.05 \\ \hline
linear & l &  & 3.07 & 11.05 & 5.39 & 12.83 \\
& r & & 3.14 & 10.69 & 5.02 & \textbf{11.78} \\
& r & ReLU  & 3.04 & 10.56 & 5.37 & 12.85 \\ 
& r & GELU  & 3.12 & 10.28 & 4.98 & 12.75 \\ \hline 
\cite{chen2020simple} & r &  & 3.11 & 10.47 & \textbf{4.67} & 12.20 \\ 
\cite{chen2020simple} & r & ReLU  & 3.04 & 10.53 & 4.92 & 12.06 \\ 
\cite{chen2020simple} &  r & GELU  & \textbf{3.02} & 10.69 & 5.52 & 13.59 \\  \hline
\cite{grill2020bootstrap} &  r & & 3.12 & 10.66 & 5.11 & 12.56 \\
\cite{grill2020bootstrap} &  r & ReLU & \underline{3.17} & \textbf{10.20} & 5.14 & \underline{14.49} \\ 
\cite{grill2020bootstrap} &  r & GELU  & 3.04 & 10.70 & 5.17 & 13.56 \\ \hline
\cite{chen2021empirical} &  r & & 3.07 & 10.92 & 5.28 & 12.67 \\
\cite{chen2021empirical} & r & ReLU & 3.05 & \underline{11.22} & \underline{5.69} & 14.45 \\ 
\cite{chen2021empirical} & r & GELU & 3.14 & 10.87 & 5.01 & 12.98 \\  \hline
\end{tabular}
\label{tab:arch}
\end{center}
\end{table}

This section ablates different aspects of network design choices when using self-supervised learning frameworks.
We investigate patch embedding and projection head design.
Table~\ref{tab:arch} reports the average rotational error on LM and LM-O for the investigated aspects.

\textbf{Projection Head}
The works of~\cite{chen2020simple,grill2020bootstrap,chen2021empirical} use high-dimensional, multi-layered projection heads to project the feature output of the backbone to the desired dimensionality.
The work of~\cite{chen2020simple} uses a two-layered MLP, with the first ReLU~\cite{nair2010rectified} and the second layer linearly activated. 
The work of~\cite{grill2020bootstrap} and~\cite{caron2021emerging} both use three-layered MLPs, yet different versions.
The latter using GELU~\cite{hendrycks2016gaussian}-activated hidden layers and weight normalization~\cite{salimans2016weight}.
In~\cite{chen2021empirical}, the projection head of~\cite{grill2020bootstrap} and the prediction head of~\cite{chen2020simple} are combined. 
Features are normalized using batch normalization~\cite{ioffe2015batch}, and hidden layers are ReLU-activated.
We compare using these projection heads to using no head or a single linear layer as head.
Since using no head requires using the backbone's output as it is, the descriptor dimensionality per feature map location is $384$.
For all the evaluated projection heads a hidden dimension of four times the output dimension of the previous stage and batch normalization are used.
We have tested with and without using weight normalization as used by~\cite{caron2021emerging}. 
Compared to batch normalization both consistently lead to increased rotational error of the matched templates.

Table~\ref{tab:arch} compares the average rotational errors of different projection heads on LM and LM-O. 
The lowest error per set is indicated in bold, the highest is indicated with an underline.
The lowest errors on seen and unseen LM, and unseen LM-O occur with heads with less layers.
Using no head leads to comparably high errors.
When using projection heads, the highest errors over all sets occur using higher dimensional heads.
In general, for the seen objects the results are similar for all heads.
Yet, projection heads with a smaller number of layers lead to less rotational error on unseen objects.
This evaluation stands in contrast to self-supervised ViTs for classification that use projection heads with $>=3$ layers and high dimensional hidden and last layers~\cite{chen2021empirical,caron2021emerging}.
The choice of activation appears to have little influence. 
Yet, heads with a lower number of layers shows reduced error on unseen objects when using no activation function.

\textbf{Patch embedding}
The authors of~\cite{chen2021empirical} propose to use random patch embedding to increase stability during training.
We experiment with the initialization of the convolution layer used for patch embedding.
The second column in Table~\ref{tab:arch} (p.e.) ablates the influence.
Updating the pre-trained patch embedding layer during fine-tuning is referred to as learned (l).
With a slight abuse of denotation we refer to not updating the patch embedding layer during fine-tuning as random (r).
We observe a similar effect as in~\cite{chen2021empirical}.
While the error difference for the seen objects is insignificant, using random patch embedding leads to significantly less error on the unseen objects.

\subsection{Self-Attention}

\begin{figure}[htbp]
\centering
\includegraphics[width=1.0\columnwidth]{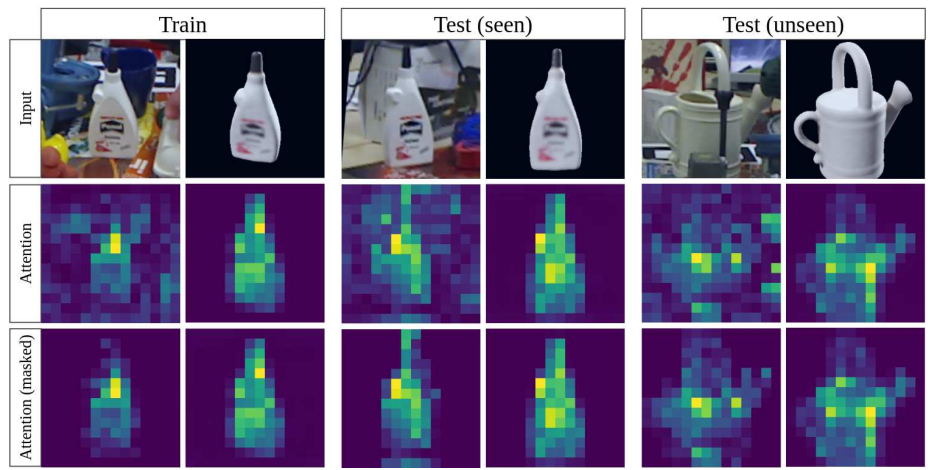}
\caption{\textbf{Self-Attention on LM/LM-O} Visualized is the self-attention of the first head of the last self-attention layer using the positional tokens as input.}
\label{fig:sa_lm}
\end{figure}

\begin{figure}[htbp]
\centering
\includegraphics[width=1.0\columnwidth]{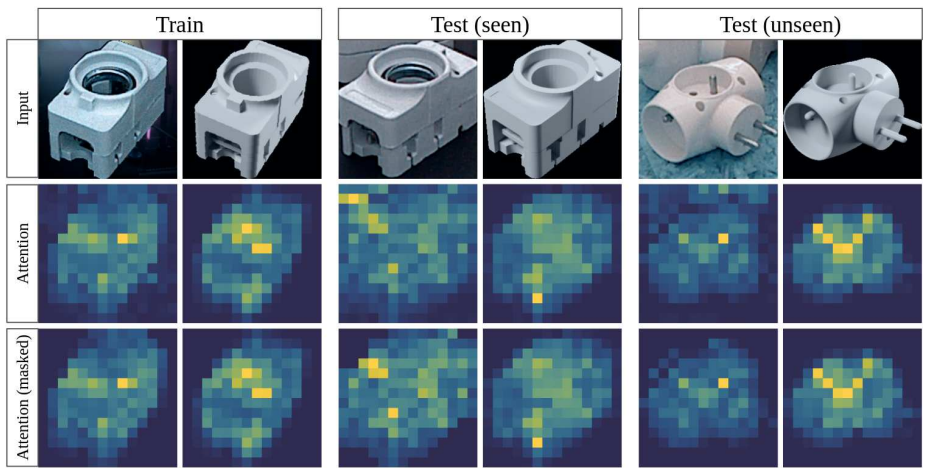}
\caption{\textbf{Self-Attention on T-LESS} Visualized is the self-attention of the first head of the last self-attention layer using the positional tokens as input.}
\label{fig:sa_tless}
\end{figure}

Figures~\ref{fig:sa_lm} and ~\ref{fig:sa_tless} visualize self-attentions maps on the training and test sets of LM/LM-O and T-LESS, respectively.
The same projection mechanism as in~\cite{caron2021emerging} is used.

On LM/LM-O, Figure~\ref{fig:sa_lm}, ViT-s effectively learns to encode relevant features of the seen objects. 
The unseen test case shows that the learned self attentions not only transfer the concept of objectness to unseen objects, but also manages to distinguish relevant from irrelevant feature map locations. 
%Due to the few templates used for training on LM, there are visible offsets between the input query image's and the template's that are matched.
%Though, the training input indicates a visible rotation offset between Even using just $301$ templates for LM

On T-Less, Figure~\ref{fig:sa_tless}, object crops often show dataset objects in front or behind the query object, as is visualized in the seen and unseen test images. 
Cropping the feature map using the template's mask is important in order to improve matching accuracy.

\section{Conclusion}
\label{sec:Conclusion}

This work presents diverse empirical analyses for using ViTs for self-supervised template matching for $3D$ pose retrieval.
The presented findings are threefold.
Using ViTs for deep template matching improves matching accuracy for seen and novel objects, in comparison to CNNs.
Using pre-trained ViTs in a zero-shot fashion, that is without fine-tuning, already exhibits strong matching accuracy. Depending on the object set and metric used for evaluation, even improving over using a similar, fine-tuned CNN-based approach.
For the problem of self-supervised synthetic template to real query object matching the network architecture is different to a comparable CNN approach and to self-supervised ViTs for image classification.
In comparison to CNNs, ViTs benefit more from pre-training due to their feature extraction being more general.
And in comparison to self-supervised ViTs for image classification, large, multi-layered projector heads are detrimental to the matching accuracy on novel objects.
We hypothesize that this occurs due to the stronger overfitting of deeper heads on the seen examples during fine-tuning, in turn harming the generality of the features learned during pre-training.
Future work will investigate how to effectively exploit the features learned during ViT pre-training.
%Further investigations will be performed with respect to the performance improvements when using ViTs for $6D$ novel object pose estimation. 

%\section{Acknowledgements}
%We gratefully acknowledge the support of the EU-program EC Horizon 2020 for Research and Innovation under grant agreement No. 101017089, project TraceBot and the NVIDIA Corporation for supporting this research by providing hardware resources.
\bibliographystyle{splncs04}
\bibliography{egbib}

\end{document}